\documentclass{article}
\usepackage{ijcai24}

\usepackage{times}
\usepackage{soul}
\usepackage{url}
\usepackage[hidelinks]{hyperref}
\usepackage[utf8]{inputenc}
\usepackage[small]{caption}
\usepackage{graphicx}
\usepackage{amsmath}
\usepackage{amsthm}
\usepackage{booktabs}
\usepackage{xcolor}
\usepackage{graphicx,subfigure}
\usepackage[switch]{lineno}

\newtheorem{definition}{Definition}
\newtheorem{lemma}{Lemma}

\newtheorem{proposition}{Proposition}

\usepackage[ruled,vlined]{algorithm2e}

\newtheorem{theorem}{Theorem}

\title{From Abstractions to Grounded Languages for \\Robust Coordination of Task Planning Robots}

\author{
    Yu Zhang
    \affiliations
    Arizona State University
    \emails
    yzhan442@asu.edu
}

\begin{document}

\maketitle

\begin{abstract}
In this paper, we consider a first step to bridge a gap in coordinating task planning robots. 
Specifically, we study the automatic construction of languages that are {\it maximally flexible} while being {\it sufficiently explicative} for coordination.
To this end,
we view language as a machinery for specifying temporal-state constraints of plans. 
Such a view enables us to reverse-engineer a language from the ground up by mapping these {\it composable} constraints to words. 
Our language expresses a plan for any given task as a ``{\it plan sketch}'' to convey just-enough details while maximizing the flexibility to realize it, leading to robust coordination with optimality guarantees among other benefits.
We formulate and analyze the problem, provide an approximate solution, 
and validate the advantages of our approach under various scenarios 
to shed light on its applications.
\end{abstract}


\section{Introduction}

To facilitate explicit coordination via communication between robots in distributed systems, 
a key consideration is the adoption of a ``{\it language}'' that the robots can all speak.
Such a language often relies on words with predefined meanings that are designed by human users~\cite{barbuceanu1995cool,xuan2001communication,finin1994kqml,poslad2007specifying}.
However, such languages tend to be either too rigid or too forgiving,
leading to brittle solutions, excess negotiation costs, or unexpected coordination issues (e.g., deadlocks). 
In this paper, as a first step, we consider to bridge the gap for task planning robots using symbolic planning. 
Traditional methods for explicit coordination via communication 
in distributed systems with planning agents can be divided into two classes:

1) {\it Centralized plan and distributed execution}: provides optimality guarantees except when approximate solutions are considered~\cite{sharon2015conflict,nissim2010general,oliehoek2016concise}. 
Note that the planning process may be centralized or distributed~\cite{nissim2010general}.
Explicit coordination in this class involves broadcasting the centralized plan in the planning language and sometimes exchanging messages as stipulated by the plan during plan execution.
This approach 
results in brittle solutions (i.e., a single agent changing its part of plan requires the entire plan to be updated) among other limitations. 

2) {\it Distributed plans and distributed execution}: provides no guarantee of optimality in general~\cite{bnaya2014conflict,zhang2016discof}.
Note that distributed plans imply that the planning process is distributed.
Methods in this class are often rule or local-search based~\cite{parker1998alliance,arai2002advances}, making them adaptive to local changes and easy to implement.
For explicit coordination, a language is often designed manually on a case-by-case basis, which is prone to unexpected coordination issues (e.g., deadlocks).

\begin{figure}[t] 
        \center{\includegraphics[width=0.40\textwidth]{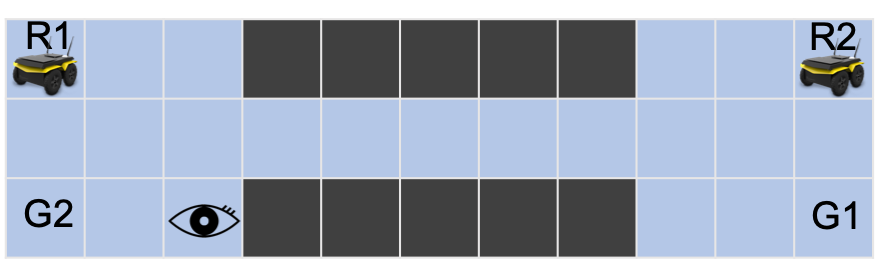}}
\caption{Motivating scenario involving two pathfinding robots, $R_1$ and $R_2$, in a gridworld. 
Each cell can only accommodate a single robot at a time and the darker cells are obstacles. 
The robots are tasked to reach their goal locations ($G_1$ and $G_2$), respectively, in the shortest timespan while avoiding collisions.
They have a limited sensing range and communication is costly. 
During plan execution, there may be locations of interest popping up at random places that require one of the robots to visit (denoted by the eye sign). 
}
\label{fig:motivatingrobot}
\vskip-15pt
\end{figure}

Our work serves as a middle ground that bridges the two classes and
combines their advantages,
contributing a novel perspective for explicit coordination between task planning agents. 
Consider the scenario in Fig. \ref{fig:motivatingrobot}. 
The problem is difficult for the second class of methods:
the robots must coordinate before one of them enters the narrow pathway so methods based on local information only would not work well (i.e., leading to deadlocks).
Furthermore, given that the locations of interest are unpredictable, neither can we assign fixed priorities to the robots (e.g., always letting $R_1$ go through first).
While these issues are not present in the first class since the robots coordinate a plan before execution, 
whenever some location of interest pops up during plan execution,
the robots must re-coordinate a plan, significantly increasing the cost.

While similar to the first class, robots in our approach coordinate by communicating  ``{\it plan sketches}''  that guarantee optimality while maximizing flexibility to reduce the need for replanning and re-coordination (thus differing from work on replanning or plan repair for making replanning more efficient~\cite{fox2006plan}). 
In the scenario above, the robots can communicate that ``{\it $R_2$ to wait for $R_1$}'' without specifying the exact plan to be followed, 
so that robot $R_2$ later (instead of $R_1$ even though $R_1$ would detect the location of interest first, since that would require $R_1$ to wait for $R_2$) can adjust locally to visit the location of interest even if its original plan does not pass through it. 
A crucial property 
is to enable
each robot to make local changes without any negative impact to the (global) makespan 
as long as the updated plan
is still consistent with the plan sketch.

To this end, we view language as a machinery for specifying plan constraints. 
A language thus specifies a {\it plan space abstraction} where a sentence in the language (i.e., a plan sketch) specifies a set of plans. 
The robots all commit to the same set of plans as a result of coordination (i.e., one robot communicates a plan sketch to the others). 
To guarantee the feasibility of this approach, 
given that the robots may be unaware of the local changes made by the others,
one of key challenges is for the plans in this set to not introduce {\it miscoordinations}; 
to maximize flexibility, the number of plans should be maximized. 
Since different sets will be specified for different tasks, we instead minimize the number of words in the language, resulting in a {\it minimal language}.  
In our approach, we associate words in the language with temporal-state constraints that are composable.
We show that searching for a minimal language 
is NEXP-complete. 
In light of this result, we develop an approximate solution.
We validate the benefits of the languages under various application scenarios.
The contribution of our work is both theoretical and practical in nature.
First, we propose a new perspective on language formation for explicitly coordinating task planning robots to bridge the two existing classes of methods 
and combine their unique advantages.  
Second, we formally analyze the language formation problem. 
Third, we provide an approximate solution by relating it to traditional planning problems. 
Finally, we provide a comprehensive evaluation of our approach with synthetic and robotic simulations.

\section{Related Work}

The emergence and evolution of language has been studied extensively in evolutionary and computational linguistics~\cite{steels2003evolving,christiansen2003language,cangelosi2012simulating}. 
Most prior work there has considered the problem in the context of evolutionary games~\cite{steels2012grounding}, and often in an iterated learning setting~\cite{kirby2014iterated}.
Steedman's work~\cite{steedman2002plans} is particularly inspiring in which he suggests a connection between natural languages and a hidden planning language that preexists in mind, although it falls short of establishing their computational connections. 
Our work introduces a framework that establishes the connection between a {\it coordination language} and the underlying planning language 
where the new language arises solely from the need for coordination. 
From a computational perspective, our work is also inspired by a prior study  
that considers the incentive of communication in a game-theoretical framework~\cite{Allott2006}.

A language often represents a structured symbolic system mapping symbols to semantic meanings that can be grounded in the environment~\cite{harnad1990symbol,steels2012grounding,tellex2011approaching,Tellex-RSS-14,gong2018temporal}.
In this work, we instead reverse engineer the process by considering the mapping from temporal-state constraints to symbols for language construction.  
These symbols (i.e., words) are then used to form sentences, which introduce a {\it  plan space abstraction} to resolve miscoordinations.  
The idea of applying abstraction to planning problems is not new. 
Most prior work has focused on state abstraction for problem decomposition,
which has been well studied in both path and task planning methods~\cite{KAMBHAMPATI1995167,kvarnstrom2000talplanner,erol1994htn}.
Such decomposition has also been shown to benefit communication and coordination in multi-agent planning~\cite{clement2007abstract,oliehoek2021sufficient}.
The temporal-state constraints for plan space abstraction used in our approach resemble options in semi-MDP and LTL expressions in temporal logic~\cite{SUTTON1999181,vardi1996automata,emerson1990temporal}.
However, these prior approaches have mostly focused on applying plan space abstraction to improving planning~\cite{abel2017near,SUTTON1999181}, or  learning such abstraction for problem decomposition~\cite{konidaris2014constructing,konidaris2012robot}.
We instead consider plan space abstraction for coordination. 
Coordination program synthesis with LTL and CSP is highly relevant but focused on addressing only a specific problem instance. 
It thus differs substantially from our work targeting a general machinery for robustness coordination under any instances, 
even though the coordination mechanism under a given instance may appear similar. 

Our problem may also appear similar to the problem of learning or planning to communicate. 
Along this line, there exists work on developing communication schemes to maximize team performance,
where the problem could be considered either in a planning or learning setting \cite{sukhbaatar2016learning,Goldman:2003,mordatch2018emergence,andreas2017translating}.
For example, a communication scheme can arise from a Dec-POMDP framework where communicating actions are modeled specifically~\cite{Goldman:2003}.
Such actions essentially augment the observation function. 
The focus there is on exploiting the additional information that can be associated with communication to maximize the expected return.
No explicit connection is established between the communication symbols and their semantic meanings (i.e., the language is not grounded). 

Prior works on task environment and agent modeling (such as TAEMS~\cite{decker1996taems} and BDI~\cite{rao1995bdi}) represent generic frameworks for formulating various domains involving interacting agents. The focus there is on environment and problem modeling, where guidelines for coordination via communication are specified but details are left to domain designers. 
In contrast, we directly address the optimization for such coordination. 
A similar situation is with ROSPlan~\cite{cashmore2015rosplan} for planner integration with execution and monitoring. Hence, these prior works are orthogonal to ours; they can benefit from our work to facilitate coordination. The work of ~\cite{nir2020automated} addressed similar coordination requirements via replanning and social rules (e.g., using {\it waiting}). 
As discussed earlier, our work bridges a gap by reducing replanning while guaranteeing optimality.

\section{Problem Formulation}

\begin{figure}[t] 
        \center{\includegraphics[width=0.35\textwidth]{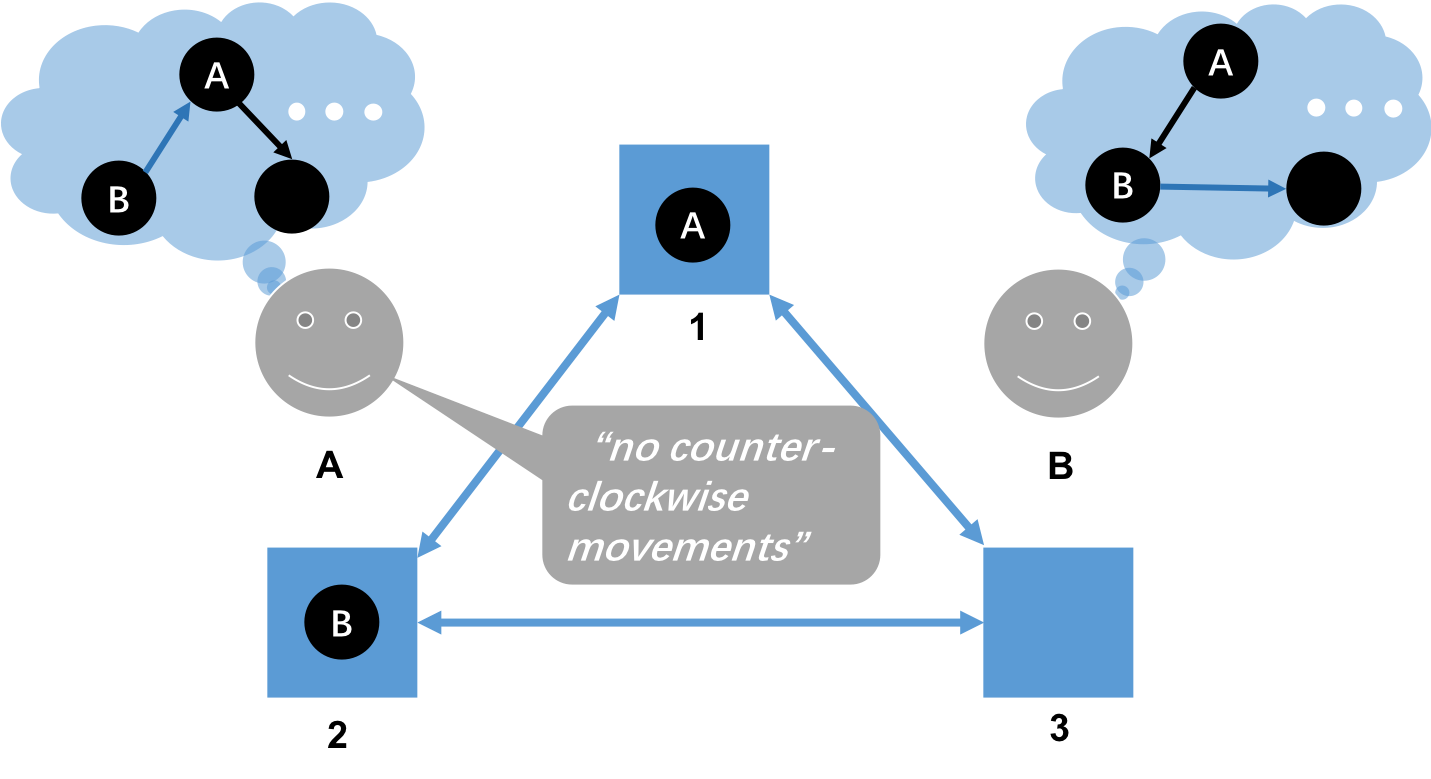}}
\vskip-10pt
\caption{{Running example where the agents cannot switch locations or stay in the same location in the same time step to avoid collisions. 
Consider a task for the two robots to switch their locations as shown. 
Without coordination, each robot may choose any candidate plan for the task and follow the corresponding subplan. 
In a situation where the plans are chosen differently as shown in the thought bubbles, 
it would lead to a miscoordination (i.e., a collision). 
}
}
\label{fig:motivating}
\vskip-10pt
\end{figure}

We adopt a joint state and action representation: whenever referring to a plan, we refer to a {\it joint and optimal} plan. 
The sequence of local actions from each robot's perspective is referred to as a {\it subplan}. 
A plan is constituted of robot subplans, which, in turn, can form into plans (details later). 
Similarly, states and actions refer to joint states and actions. 
Substate and subaction refer to each robot's local state and action.
Proofs are included in the supplementary.

\subsection{{Problem Setting}}
We focus our discussion on the problem setting that involves only two robots with the following assumptions:
\begin{itemize}
	\item $A1$: Both robots have access to the full planning domain model, the task information, and sufficient computational resources, such that either robot can compute plans independently. 
	\item $A2$: Each robot may not observe the other robot or changes to the environment made by the other during plan execution (so as to focus on explicit coordination).
\end{itemize}

$A1$ results in a special case of cooperative multi-agent planning setting where agents maintain accurate beliefs about the domain models of the others~\cite{gmytrasiewicz2005framework}. 
See Sec. \ref{sec:conclusion} about relaxing these assumptions. 
Without coordination, each robot may choose any candidate plan for the task and execute the corresponding subplan.
Coordination is needed to ensure that the robots do not choose plans that lead to {\it miscoordinations},
i.e., situations when choosing different plans leads to suboptimality or task failures. 
To allow local changes (i.e., changes to subplans) that are critical for flexibility, 
part of the aim is to retain as many candidate plans as possible. 
However, more candidate plans will more likely lead to miscoordinations, creating a challenging tradeoff.  
Intuitively, the solution is to have the robots both commit to the same set of plans (a subset of all candidate plans for the task) via coordination. 
The challenge that remains is to identify the largest sets of plans for different tasks where local changes can be {\it independently} made without affecting the (global) task performance. 

\subsection{{Running Example}}

To facilitate the discussion, we further introduce a running example in Fig. \ref{fig:motivating} with two robots $\{A, B\}$ and three locations $\{1$, $2$, $3\}$.
We consider a coordination process where either robot communicates a ``{\it plan sketch}'' (specifying a set of candidate plans) for both robots to commit to. 
A plan sketch may correspond to a set of plans having certain properties.
For example, $A$ can tell $B$ to {\it not} move counter-clockwise, which resolves the miscoordination illustrated in Fig. \ref{fig:motivating}. 
In contrast to $A$ directly communicating its plan to $B$, this way, $B$ has the flexibility to either wait first and then move, or move first and then wait,
resulting in more robust coordination. 


\subsection{{Preliminaries}}
We model the planning domain based on a slightly modified STRIPS model~\cite{FIKES1971189} as $M = (P, A)$,
where $P$ is the set of propositional state variables 
and $A$ is the set of actions.
Each action $a \in A$ specifies a pair of robot subactions such that $a = \langle a^A, a^B \rangle$, where $\langle \rangle$ denotes an ordered set.
Each action $a$ is associated with a set of preconditions, $pre(a) \subseteq P$, add effects, $add(a) \subseteq P$,  
and delete effects, $del(a) \subseteq P$.
We consider a set of candidate tasks in the domain for our language formation problem,
which represents all relevant tasks to the robots. 
Each candidate task is in the form of a pair $(I, G)$, where $I \in 2^P$ and $G$ specifies the propositions required to be present in the goal state. 
Each $(I, G)$ pair introduces a planning problem $O = (P, A, I, G)$.
A plan $\pi$ is a sequence of actions:
\begin{equation}
\pi = \langle \langle a_1^A, a_1^B \rangle .. \langle a_n^A, a_n^B \rangle \rangle
\end{equation}
where $n$ is the length of the plan. 
$\pi$ may also be specified as the combination of two subplans as $\pi = \langle \pi^A, \pi^B\rangle$,
where $\pi^A = \langle a_1^A,.., a_n^A\rangle$ denotes the subplan for robot $A$
and $\langle \rangle$ here may also be viewed as an operator that element-wisely combines $\pi^A$ and  $\pi^B$ into a plan.
When combined, $\pi^A$ and $\pi^B$ are assumed to be aligned from the first step and onward,
and the shorter subplan is padded with {\it idle} subactions.

Given a domain model $M$, the resulting state after executing plan $\pi$ in state $s$ is determined by the transition function $\gamma$, which is defined as follows where `$\cdot$' denotes concatenation:
\begin{equation}
\gamma (\pi, s)=
\begin{cases}
  s & \text{if }
       \!\begin{aligned}[t]
       \pi = \langle\rangle \\
       \end{aligned}
\\
  \gamma(\langle a\rangle, \gamma(\pi', s))  & \!\begin{aligned}[t]
       \text{else } \pi = \pi' \cdot \langle a \rangle \\
       \end{aligned}
\end{cases}
\end{equation}

The transition function $\gamma$ for an action sequence with a single action $a$ and state $s$ is defined as:
\begin{equation}
\gamma (\langle a \rangle, s)=
\begin{cases}
  ( s \setminus Del(a)) \cup Add(a) & \text{if }
       \!\begin{aligned}[t]
       Pre(a) \subseteq s \\
       \end{aligned}
\\
  s & \text{otherwise}
\end{cases}
\end{equation}

Given a planning problem $O = (P, A, I, G)$,
an action sequence $\pi$ is a plan for $O$
 iff $\gamma (\pi, I)$ entails $G$ {\it and} $\pi$ has the minimal cost.
 For simplicity, we assume that every action has a cost of $1$ and hence a plan for $O$ minimizes the makespan to satisfy $G$ from $I$. 
A plan $\pi$ also introduces a state sequence, denoted by $\pi_S$,
 such that the first element $\pi_S[0] = I$ and the last $\pi_S[n] = \gamma (\pi, I)$ entails $G$ with a plan length $|\pi| = n$.
 

\subsection{{Required Coordination}}
Next, we formally define {\it required coordination} that introduces miscoordinations.
Under our problem setting, a miscoordination may occur when robots choose different candidate plans. 
Given a planning problem $O = (P, A, I, G)$, 
$\Pi(O)$ denotes the set of all candidate plans for problem $O$.
Without coordination, either robot can choose any plan in $\Pi(O)$.

\begin{definition}[Required Coordination (RC)]
Given a planning problem $O = (P, A, I, G)$, 
a required coordination is the following condition: $\exists \pi_1, \pi_2 \in \Pi(O)$ $({\pi_1 \neq \pi_2})$, 
$\langle \pi_1^A, \pi_2^B\rangle \not\in \Pi(O)$ 
$|$ $\langle \pi^A_2, \pi^B_1\rangle \not\in \Pi(O)$,
where $\pi_1 = \langle\pi_1^A, \pi_1^B\rangle, \pi_2 = \langle\pi_2^A, \pi_2^B\rangle$. 
\label{def:rc}
\end{definition}

\begin{figure}[t]
        \center{\includegraphics[width=0.42\textwidth]{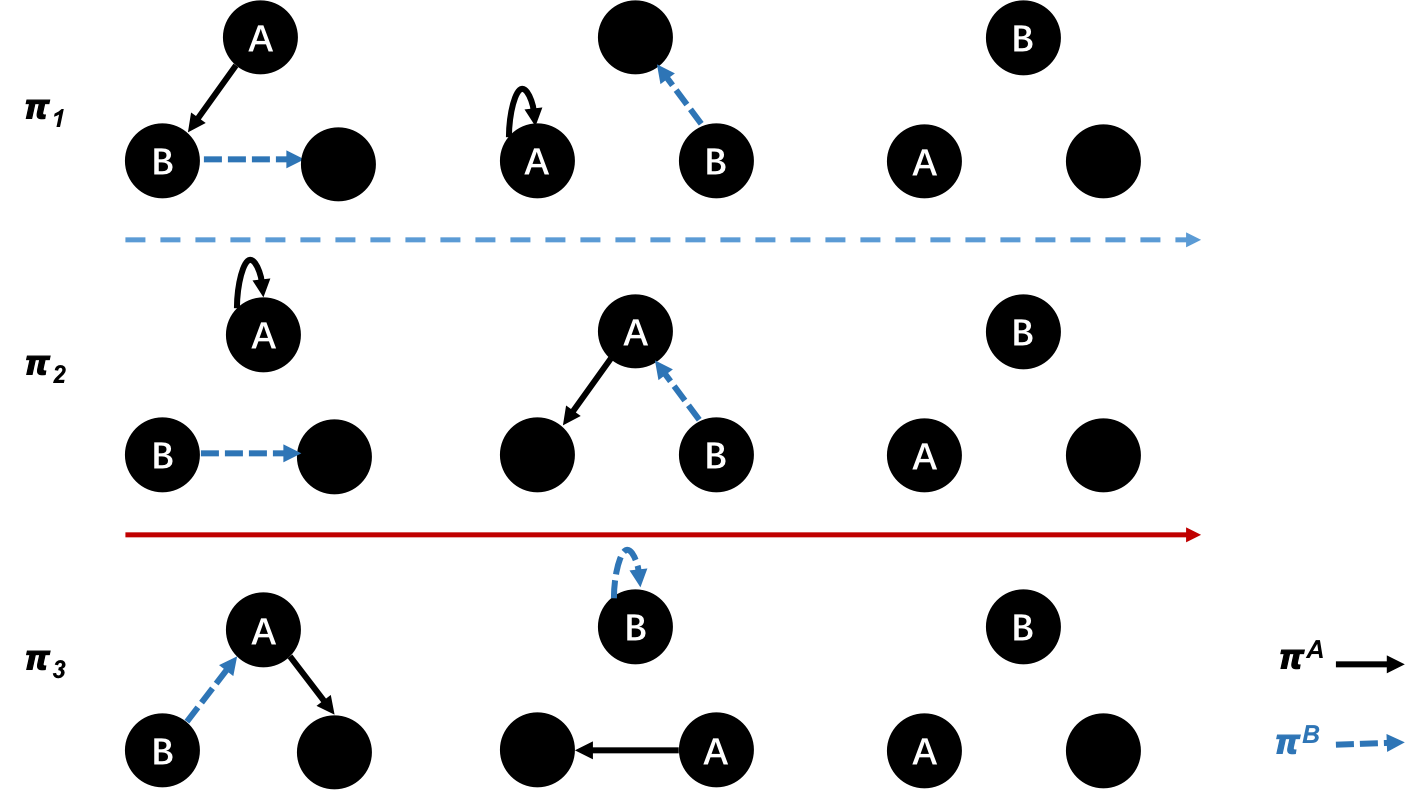}}
\caption{{Three different plans for the example in Fig. \ref{fig:motivating}:  the first two plans do not introduce RC: subplan $\pi_1^A$ (black arrows) in $\pi_1$  can be switched with its counterpart ($\pi_2^A$) in $\pi_2$ and recombined with subplan $\pi_2^B$ (blue dashed arrows) 
without introducing any miscoordinations. 
Both $\pi_1$ and $\pi_2$, however, introduce RC with $\pi_3$.}}
\label{fig:failure}
\vskip-10pt
\end{figure}

Intuitively, an RC condition defines a situation where there are two different plans $\pi_1, \pi_2 \in \Pi(O)$ for problem $O$ 
but the recombination of the subplans does not belong to $\Pi(O)$ (i.e., not a plan for $O$). 
If the set of candidate plans that the robots commit to includes these two plans, a miscoordination may occur when the robots choose $\pi_1$ and $\pi_2$, respectively.
In such cases, we say that the two plans {\textit{{introduce}}} RC. 
Note that two different plans do not necessarily introduce RC (Fig. \ref{fig:failure}).

\begin{proposition}
Given a domain model $M = (P, A)$ and a set of candidate tasks $\mathcal{T}$,
coordination is necessary if and only if the following holds:  $\exists \pi_1, \pi_2 \in \Pi(O)$,  $(\pi_1, \pi_2)$ introduces RC, where $(I, G) \in \mathcal{T}$ and $O = (P, A, I, G)$. 
\label{th:rc}
\end{proposition}

Given any planning problem $O$ above,
if no plan pair exists that introduces RC, we consider the following two possible cases.
$1)$ No plan or only a single plan exists: in such a case, 
either the robots will both fail to find a plan or they will find the same plan. 
No coordination is necessary.
$2)$ There are multiple plans but none of them pair-wisely introduce RC. 
In such a case, either robot can select any candidate plan for $O$, 
and the recombined plans are guaranteed to be in $\Pi(O)$ (Def. \ref{def:rc}) and it is hence miscoordination-free.

In Fig. \ref{fig:failure}, 
$\pi_1$ and $\pi_2$ do not introduce RC. 
Hence, had these two plans being the only plans in the set of candidate plans that the robots commit to (as a result of coordination) for a task, 
no additional coordination would be necessary.
This observation already hints on our language construction. 
In particular, a language will 
be required to {\it always} express $\{\pi_1, \pi_2\}$ and $\{\pi_3\}$ differently under the task in Fig. \ref{fig:motivating}.

\subsection{Coordination Language}

The ability to separately express different sets of candidate plans requires a language to specify constraints on plans, essentially forming plan state abstraction. 
Since plans are temporal state sequences, we consider words as symbols that map to temporal-state constraints.
These symbols can be further combined to form sentences for expressing more complex constraints.
We use $:$ to indicate a range of indices (inclusive at both ends).
For example, $\pi_S[x:y]$ returns the set of states in $\pi_S$ indexed from $x$ to $y$.
An {\it abstract state} $S_i$ corresponds to a subset of the state space $S$  (i.e., $S_i \subseteq S$).

\begin{definition}[Temporal-State Constraint (TSC)]
A TSC over a state space $S$ specifies a constraint $\zeta$ in the form of
$\zeta = {S_1 \dots S_i \dots S_m}$, where $S_i \subseteq S$, or $S_i$ is an abstract state. 
\end{definition}

\begin{figure}
\centering
{
    \includegraphics[width=0.40\textwidth]{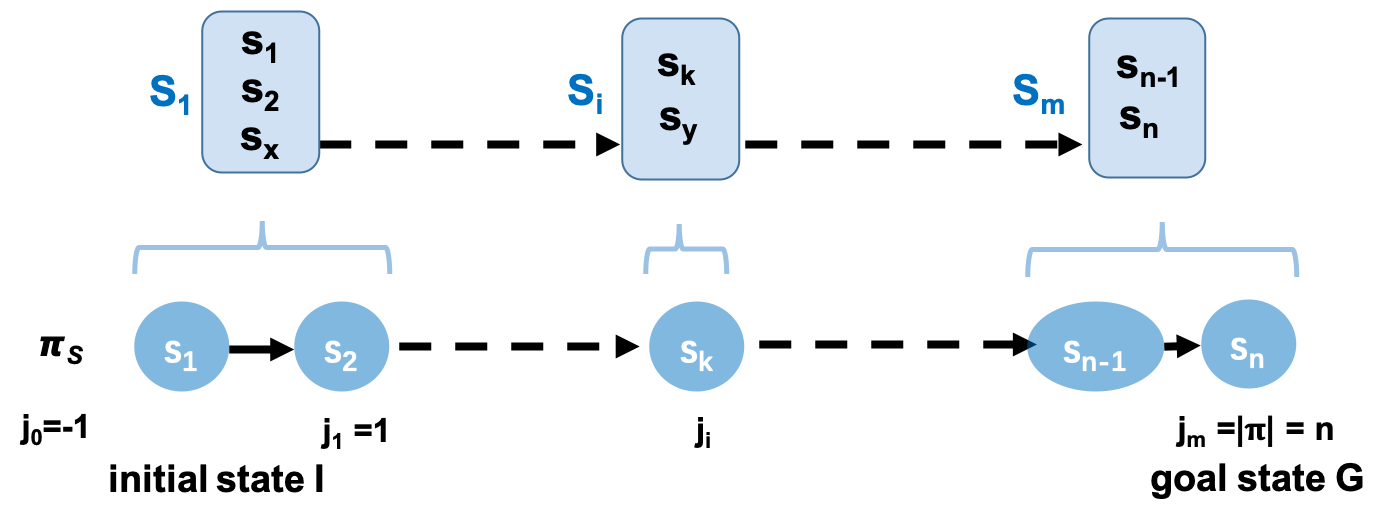}
}
\caption{{An illustration of a TSC. The top shows the abstract states associated with the TSC and the bottom a plan as a state sequence that satisfies it. The rectangle nodes represent abstract states with their included ground states shown inside.}}
\vskip-10pt
\label{fig:absplan}
\end{figure}

A plan $\pi$ under the same $S$ {\it satisfies} a TSC $\zeta$ if 
there exists a set of strictly monotonically increasing integers $j_0, j_1, j_2 \dots j_m$ with $j_0 = -1$ and $j_m = |\pi|$,
such that $\pi_S[j_{i - 1} + 1: j_{i}] \subseteq S_i$ ($n \geq m \geq i \geq 1$).
Intuitively, a TSC breaks a state sequence into segments such that 
the states in each segment belong to the same abstract state.
Notice that a plan may satisfy multiple TSCs if the abstract states are allowed to overlap;
otherwise, the TSC for a plan will be unique. 
An illustration of a TSC is provided in Fig. \ref{fig:absplan}. 
Alternatively, we also refer to that a TSC $\zeta$ {{\textit{expresses}}} a plan $\pi$ (denoted by $\zeta \sqsupseteq \pi$), if $\pi$ satisfies $\zeta$. 
One may use more expressive constraints for language formation, such as specified by LTL and CTL~\cite{vardi1996automata,emerson1990temporal}.
We choose TSCs in this work since they are relatively easier to analyze
while still being extensible.

\begin{definition}[Language]
A language for a domain $M = (P, A)$ 
is a pair $\mathcal{L} = (\mathcal{W}, \mathcal{O})$, where $\mathcal{W}$ is the vocabulary 
and $\mathcal{O}$ is the set of compositional operators that can be applied to connect the words.
Each word $w \in \mathcal{W}$ is a TSC over $S = 2^P$ and TSCs must remain closed under $\mathcal{O}$.
\end{definition}

Operators are applied to connect the words in a language to form more complex TSCs, which are also referred to as {\textit{{sentences}}}.
A given language thus introduces a rich set of TSCs for the domain. 
Not all languages are of interest to the robots.
To be useful, a language must be able to relax RCs for any candidate tasks (following Proposition \ref{th:rc}). 
Any language with such a capability is referred to as a {\it coordination language}.
In the following, with a slight abuse of notation, we denote any TSC $\zeta$ that can be expressed by a language $\mathcal{L}$ as $\zeta \in \mathcal{L}$.

\begin{definition}[Coordination Language]
A language $\mathcal{L}$ for a domain $M = (P, A)$ is a coordination language under a set of candidate tasks $\mathcal{T}$ if the following condition holds: $\forall \pi \in \Pi(O)$ where $O = (P, A, I, G)$ and $(I, G) \in \mathcal{T}$, 
1) $\exists \zeta \in \mathcal{L}, \zeta  \sqsupseteq \pi$, 
and 2) the subset of plans in $\Pi(O)$ that are expressed by $\zeta$ must be RC-free,
or more formally, $\{\pi | \zeta \sqsupseteq \pi \land \pi \in \Pi(O)\}$ pair-wisely introduce no RC.
\label{def:ela}
\end{definition}

More intuitively, since a sentence may express multiple plans,
a coordination language ensures that, under any candidate task, 1) any plan is expressible by a sentence in the language ({\it completeness}), and 2) the plans that are expressed by any such sentence do not pair-wisely introduce RC ({\it RC-free}).
Hence, the language can be used to coordinate the robots under any candidate task. 
Finding a coordination language is not difficult.
For example, when we use words for grounded actions and concatenate them,
we can express any plan exactly. 
Such a language, however, is undesirable since it is too rigid for coordination. 
For more flexibility, we would like to maximize the number of plans expressed. 
Since the situation may differ from task to task, 
instead, 
we consider the problem to minimize the size of the vocabulary,
which also aligns with language design principles~\cite{chomsky2005three,chomsky2014minimalist,steels2011modeling}. 
To simplify its formal analysis, 
we only consider concatenation ($\cdot$)
and assume words formed by abstract states only (i.e., $\mathcal{W} \subseteq 2^S$),
which introduce a {\it state abstraction}. 
Note that TSCs are naturally closed under $\cdot$.

\begin{definition}[State Abstraction]
A state abstraction is a set of abstract states, $Z = \{S_z\}$, over a subset $S'$ of the state space $S$, corresponding to a
many-to-many mapping  $f: S' \rightarrow Z$, where a state mapping to an abstract state belongs to that abstract state.
\label{def:sabs}
\end{definition}
 

\begin{theorem}
The problem  
of deciding whether a coordination language  for domain $M$ under a set of candidate task $\mathcal{T}$ exists with vocabulary $\mathcal{W}$ as a state abstraction of a minimal size $K$ 
and $\mathcal{O} = \{\cdot\}$ (concatenation), or denoted by ${D_{MinW}} = (M, \mathcal{T}, K)$, is NEXP-complete. 
\label{NEXP}
\end{theorem}

\subsection{Finding Coordination Languages}

Next, we focus on developing an approximate solution for ${D_{MinW}}$.
We define a {\textit{{perfect state abstraction}}} as a state abstraction that introduces a partition of the state space: 
 every state belongs to one and only one abstract state. 
 $f$ in Def. \ref{def:sabs} becomes a surjective mapping over the entire state space $S$. 

\begin{figure}
\centering
{
    \includegraphics[width=0.40\textwidth]{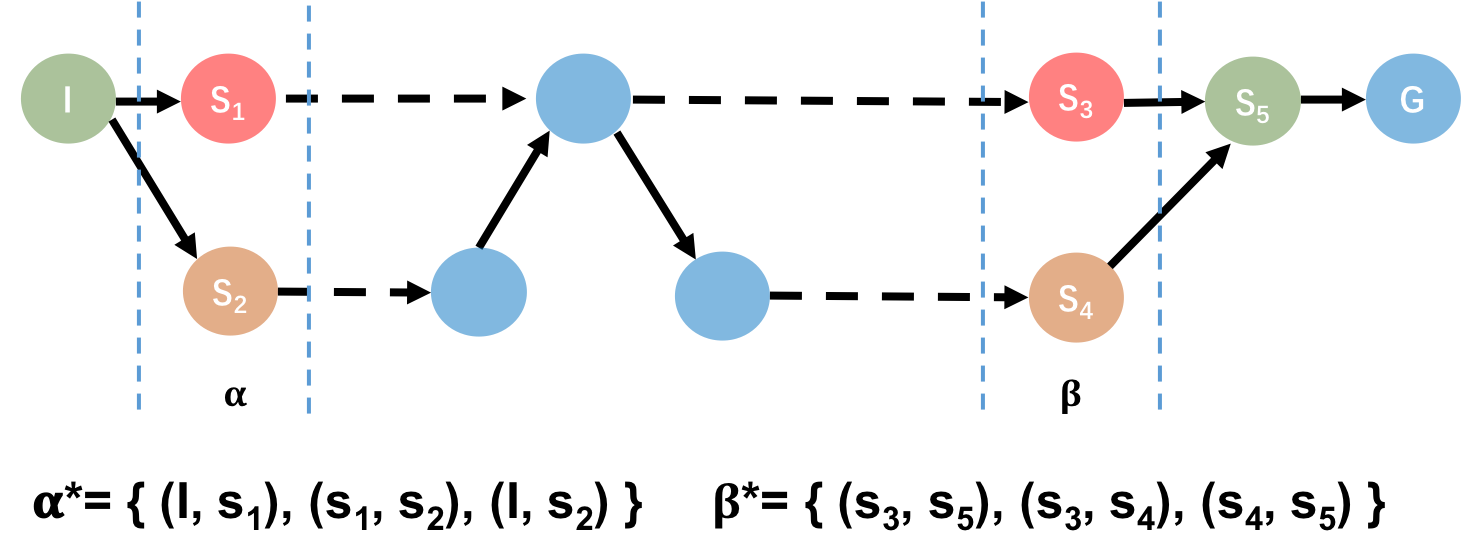}
}
\caption{{Determining $(\alpha^*, \beta^*)$ with two plans (top and bottom) that introduce RC for a given task. When \{$I$, $s_1$, $s_2$\} (or \{$s_3$, $s_4$, $s_5$\}) are pair-wisely separated into different abstract states, the two sentences for expressing these two plans will be different since they must represent $s_1$ and $s_2$ (or $s_3$ and $s_4$) as different abstract states with the preceding (or trailing) abstract state shared between the sentences, thus relaxing the RC.}}
\label{fig:d}
\vskip-10pt
\end{figure}

\begin{lemma}
Given a coordination language $(\mathcal{W}, \{\cdot\})$ for domain $M$ under a set of candidate tasks $\mathcal{T}$,
there is another coordination language, $(\mathcal{W}', \{\cdot\})$, where $\mathcal{W}'$ is perfect. 
\label{lem:replace}
\end{lemma}

\begin{theorem}
The decision problem of ${D_{MinW}}$ with a perfect vocabulary, denoted by ${D_{MinW^+}}$,  is NEXP-complete.
\label{thm:plus}
\end{theorem}

The proof in Theorem \ref{NEXP} can be applied here with minor modifications.
To simplify notations, we also use ${D_{MinW^+}}$ to denote the problem of searching for a coordination language with a minimal perfect state abstraction when there is no ambiguity.
Note that a language that uses a perfect state abstraction 
may have a reduced flexibility
since it tends to create more stringent TSCs. 
Using a perfect state abstraction implies that the TSC for expressing any plan is {\textit{{unique}}}.

For each candidate task $(I, G) \in \mathcal{T}$ of a domain $M = (P, V)$, we can compute ${\Pi}(O)$ for the induced planning problem $O = (P, V, I, G)$. 
For any plan pair $(\pi_1, \pi_2)$ in $\Pi(O)$ that introduces RC, 
we denote the first and last places where they differ as a pair $d = (\alpha, \beta)$. 
$\alpha$ and $\beta$ can be the same; $\beta$ does not always exist since the ground goal states for $\pi_1$ and $\pi_2$ may be different.
$\alpha$ ($\beta$) includes a state from $\pi_1$ and a state from $\pi_2$.
We use $\alpha^*$ ($\beta^*$) to denote the three pairs of states that are formed pair-wisely by $\alpha$ ($\beta$) with the previous (next) state shared by the plans.
See Fig. \ref{fig:d} for an illustration of $\alpha^*$ and $\beta^*$ with two plans. 
We collect $(\alpha^*, \beta^*)$ for all $(\pi_1, \pi_2)$ and for all $(I, G) \in \mathcal{T}$ as $\mathcal{D}$. 
\vskip-10pt
\begin{algorithm}
\caption{ Approx. minimal language for ${D_{MinW^+}}$}
\SetKwInOut{Input}{input}
\SetKwInOut{Output}{output}
\Input{$M, \mathcal{T}, \mathcal{D} = \emptyset$}
\ForAll{(I, G) $\in$ $\mathcal{T}$}
{
    Formulate $O = (M, I, G)$ \\
    Compute $\Pi(O)$ \\
    \ForAll{$\pi_1, \pi_2$ $\in$ $\Pi(O)$}
    {
        \If{RC is introduced} 
        {
            Determine $\alpha^*$ and $\beta^*$\\
            Add $(\alpha^*, \beta^*)$ into $\mathcal{D}$
        }
    }
 }
 Find an $f^+$ that satisfies the condition in Theorem \ref{expressive}\\
 Extract the perfect state abstraction $\mathcal{W}$ from $f^+$\\
 \Return{\{$(\mathcal{W}, \{.\})$\}}
\label{alg:alg}
\end{algorithm}
\vspace{-3mm}

\begin{theorem}
Given the $\mathcal{D}$ above induced from a given domain $M$ and a set of candidate tasks $\mathcal{T}$, if a perfect state abstraction $\mathcal{W}$ (with its mapping denoted by $f^+$) satisfies the condition that
 $\forall (\alpha^*, \beta^*) \in \mathcal{D}$, $\forall i$ $f^+(\alpha^*[i][0]) \neq f^+(\alpha^*[i][1])$ or $\forall i$ $f^+(\beta^*[i][0]) \neq f^+(\beta^*[i][1]) (i \in \{0, 1, 2\})$,
 the language of $(\mathcal{W}, \{\cdot\})$ is a coordination language for $M$ under $\mathcal{T}$. 
 \label{expressive}
\end{theorem}
Essentially, the condition above requires the three states 
that are associated with either $\alpha^*$ or $\beta^*$ between any two plans 
(that introduce RC) to be pair-wisely separated 
in the state abstraction. 
Intuitively, this enables the language to always distinguish between the two plans 
so that the associated RC can be relaxed. 
Using Theorem \ref{expressive}, we can convert the problem of searching for a minimal coordination language to the problem of finding a minimal perfect state abstraction that satisfies the condition therein. 
The algorithm is 
 in Alg. \ref{alg:alg}.



\begin{figure}
\centering
{
    \includegraphics[width=0.37\textwidth]{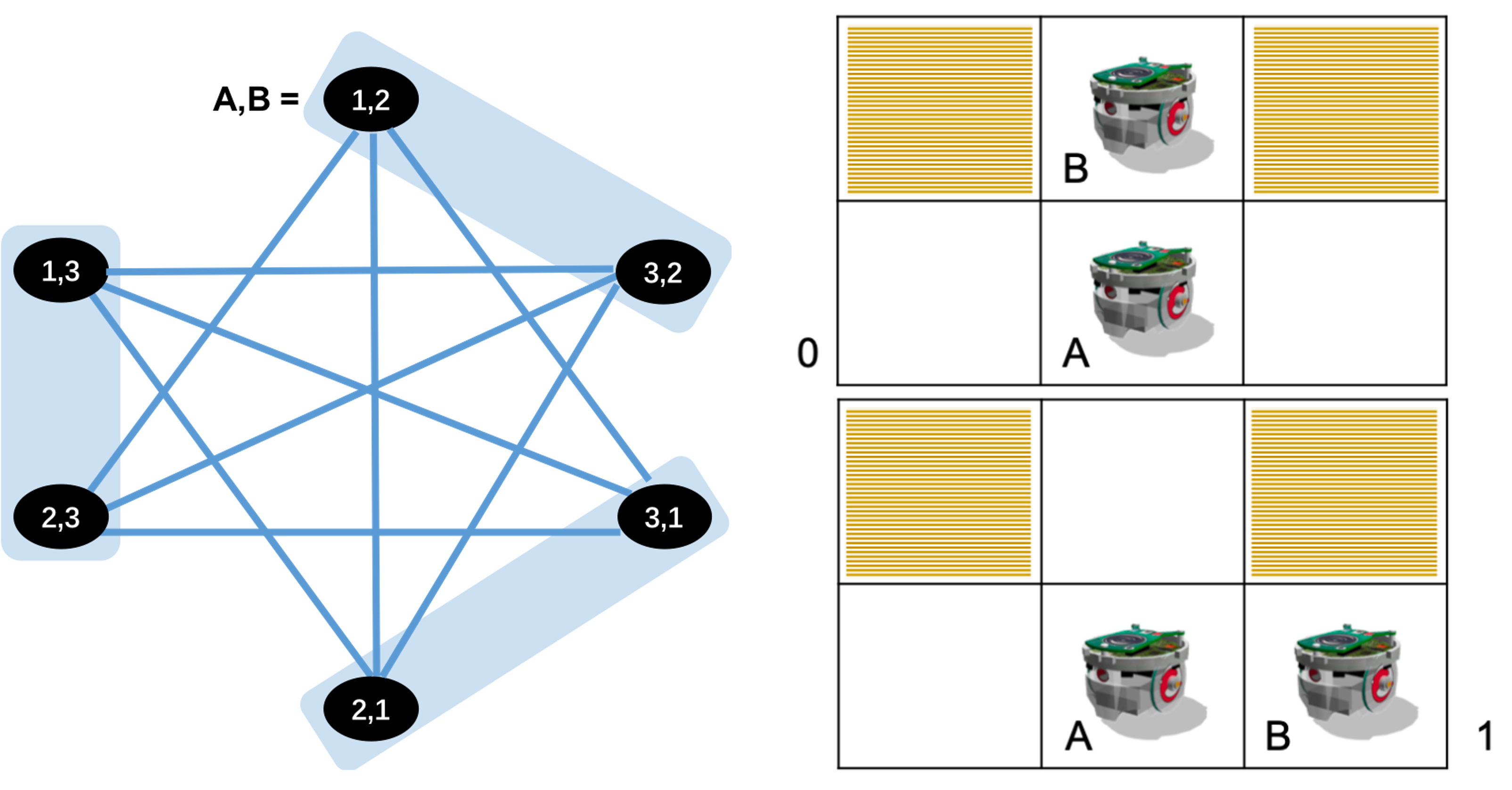}
}
\vskip-10pt
\caption{{\textbf{Left}: A possible set of state pairs (connected by edges) required to be separated by a perfect state abstraction in a coordination language for our running example, as determined by the condition in Theorem \ref{expressive}. 
Each node is a state with the first and second numbers representing the locations of $A$ and $B$, respectively.
The blue clusters present a possible vocabulary for the language. \textbf{Right}: Brown cells are non-traversable spaces. Two of the three abstract states (labeled by `$0$' and `$1$' above) for this domain contain a single ground state. The remaining ground states all belong to the third (`$2$').}}
\vskip-10pt
\label{fig:rc}
\end{figure}

The condition described above in Theorem \ref{expressive}, however, is only a sufficient condition for a coordination language--there may exist a coordination language 
that does not satisfy it. 
This implies that finding the minimal perfect state abstraction that satisfies the condition induces only an approximate solution for ${D_{MinW^+}}$. 
However, the completeness and RC-freeness of the language are unaffected. 
The positive side is that this solution only requires solving a set of planning problems (of size $|\mathcal{T}|$ to be precise),
which have efficient existing solutions.  
 Existing planning methods that return all plans, such as m-A*~\cite{dechter2012search} or DFBB~\cite{lawler1966branch}, can be used.  
For our running example in Fig. \ref{fig:motivating}, a perfect state abstraction produced by Alg. \ref{alg:alg} is presented in Fig. \ref{fig:rc} (left).
Note that this example is a special case where the longest plan length is $2$, implying that $\alpha$ and $\beta$ will always be the same and hence the 
solution returned is the exact minimal language for ${D_{MinW^+}}$.
Readers may have observed that the words in Fig. \ref{fig:rc} (left) corresponds to specifying the location of robot $B$.
The robots can use these words to compose sentences to specify clockwise or counterclockwise movements for $B$, 
consistent with our earlier discussion of Fig. \ref{fig:motivating}.

\vskip5pt
\subsubsection*{Notes on using a coordination language}
Using a coordination language requires both robots to first ``learn'' the language,
similar to how people learn the same language to communicate~\cite{gentner2006analogical}.
Learning a coordination language involves the maintenance of the language specifications (i.e., $(\mathcal{W}, \mathcal{O})$), such as the mapping from ground states to words (e.g., $f^+$).
A language only needs to be precomputed once and shared among all speaking robots. 
It can then be used by them for any candidate task. %

\section{Evaluation Results}
\label{sec:results}

We perform evaluation of our approach using both synthetic and robotic simulations under various application scenarios 
to illustrate the benefits of the proposed language. 
All scenarios are in a gridworld-based (discrete) setting, and similar to the running example. 
The candidate tasks include all possible tasks in the domain unless stated otherwise.
Simulations were created using Webots.
Details are reported in supplementary. 

\subsection{Interpreting the Generated Languages}
\label{subsec:interpreting}

First, we inspect the language in a relatively simple environment in Fig. \ref{fig:rc} (right).
We assume that the robots can only observe each other in adjacent cells
so miscoordinations can lead to collisions. 
We implemented a brute-force method that exactly computed the minimal language for ${D_{MinW^+}}$ (Theorem \ref{thm:plus}). 
The brute-force method checks through all possible perfect state abstractions 
and returns the one with the minimal size that introduces a coordination language. 
It starts with size $0$ and gradually increases it.  
The abstraction returned by the algorithm contains $3$ abstract states with
two of them containing only a single ground state in Fig. \ref{fig:rc} (right).


To make sense of the language, first, observe that 
miscoordinations can only occur in tasks where the robots must swap their locations while starting non-adjacent to each other. 
In these tasks, miscoordinations can occur  where the robots both choose to start first.
For all the other tasks, the plan is always unique such that both robots would choose the same plan so no coordination is needed.  
Hence, a coordination language needs only to specify which robot starts first.


Now, consider the 3-word language above: two words in Fig. \ref{fig:rc} (right) denoted by `$0$' and `$1$', respectively, with the third denoted by `$2$'. 
Consider a scenario where robot $A$ starts at the top middle and $B$ bottom left, and must swap locations. 
If robot $A$ is to move first, either robot can communicate ``${202}$''; similarly, for $B$ to move first, ``${212}$'' can be communicated.
Since the two plan sketches are expressed differently in the coordination language,
miscoordination is avoided.  
What is interesting here is that the ``semantic meanings'' of abstract states or words are {\it task context dependent}. 
For example, depending on the task, 
`$0$' may be used in different sentences to express either $A$ or $B$ moving first.

\subsection{Language Properties}

\begin{table}
\resizebox{8.6cm}{!}{
\begin{tabular}{|r|r|r|r|r|r|r|r|r|r|r|}
\hline
& \multicolumn{2}{|c|}{env. size} & \multicolumn{1}{|c|}{$|\mathcal{T}|$} & \multicolumn{1}{|c|}{$|S|$}  & \multicolumn{1}{|c|}{$|Z|$} & \multicolumn{1}{|c|}{time (s)} & \multicolumn{1}{|c|}{$|\approx Z|$} &  \multicolumn{1}{|c|}{time (s)} \\ \hline
GW \#1 & \multicolumn{2}{|c|}{2 $\times$ 2}& 132 & 12 & 3 & 26.2 & {\textbf{7}} & 0.1 \\ \hline
GW \#2 & \multicolumn{2}{|c|}{2 $\times$ 3}& 870  & 30 & \hrulefill & $\gg$ 3600.0 & {\textbf{13}} & 0.9\\ \hline
GW \#3 & \multicolumn{2}{|c|}{2 $\times$ 4}& 3080 & 56 & \hrulefill & $\gg$ 3600.0 & {\textbf{18}} & 19.5 \\ \hline
GW \#4 & \multicolumn{2}{|c|}{3 $\times$ 3}& 5112  & 72 & \hrulefill & $\gg$ 3600.0 & {\textbf{22}} & 86.2\\ \hline \hline
& \multicolumn{2}{|c|}{env. size} & \multicolumn{1}{|c|}{$|\mathcal{T}|$} & \multicolumn{1}{|c|}{$|S|$}  & \multicolumn{2}{|c|}{$Manhattan(I, G)$} & \multicolumn{1}{|c|}{$\approx Z$} &  \multicolumn{1}{|c|}{time (s)} \\ \hline
GWB \#1 & \multicolumn{2}{|c|}{2 $\times$ 4}& 3080     & 56    & \multicolumn{2}{|c|}{all possible tasks} & {\textbf{10}} & 100.6\\ \hline
GWB \#2 & \multicolumn{2}{|c|}{3 $\times$ 3}& 3080  & 56 & \multicolumn{2}{|c|}{all possible tasks} & {\textbf{11}} & 92.9\\ \hline
GWB \#3 & \multicolumn{2}{|c|}{3 $\times$ 4}& 8010   & 90 & \multicolumn{2}{|c|}{all possible tasks} & {\textbf{12}} & 18890.9\\ \hline
GWB \#4 & \multicolumn{2}{|c|}{2 $\times$ 5}& 8010  & 90 & \multicolumn{2}{|c|}{all possible tasks} & {\textbf{13}} & 19557.3\\ \hline
GWB \#5 & \multicolumn{2}{|c|}{3 $\times$ 3}& 380  & 56 & \multicolumn{2}{|c|}{$\geq 4$} & {\textbf{4}} & 1.4\\ \hline
GWB \#6 & \multicolumn{2}{|c|}{3 $\times$ 4}& 636  & 90 & \multicolumn{2}{|c|}{$\geq 5$} & {\textbf{4}} & 11.5\\ \hline
GWB \#7 & \multicolumn{2}{|c|}{3 $\times$ 5}& 956  & 132 & \multicolumn{2}{|c|}{$\geq 6$} & {\textbf{4}} & 132.5\\ \hline
GWB \#8 & \multicolumn{2}{|c|}{4 $\times$ 4}& 956  & 132 & \multicolumn{2}{|c|}{$\geq 6$} & {\textbf{4}} & 140.3\\ \hline
\end{tabular}
}
\caption{{Performance of the approximate solution in Alg. \ref{alg:alg} compared to a brute-force solution for ${D_{MinW^+}}$ in a multi-robot pathfinding domain. Column `$|Z|$' is the vocabulary size returned by brute-force and `$|\approx Z|$' is the size returned by the approximate solution.
The time for computing plans for all candidate tasks is included for the approximate solution.}
}
\vskip-10pt
\label{tab:abs}
\end{table}

\begin{figure*}[t!]
    \centering
    \begin{subfigure}
        \centering
        \includegraphics[width=0.325\textwidth]{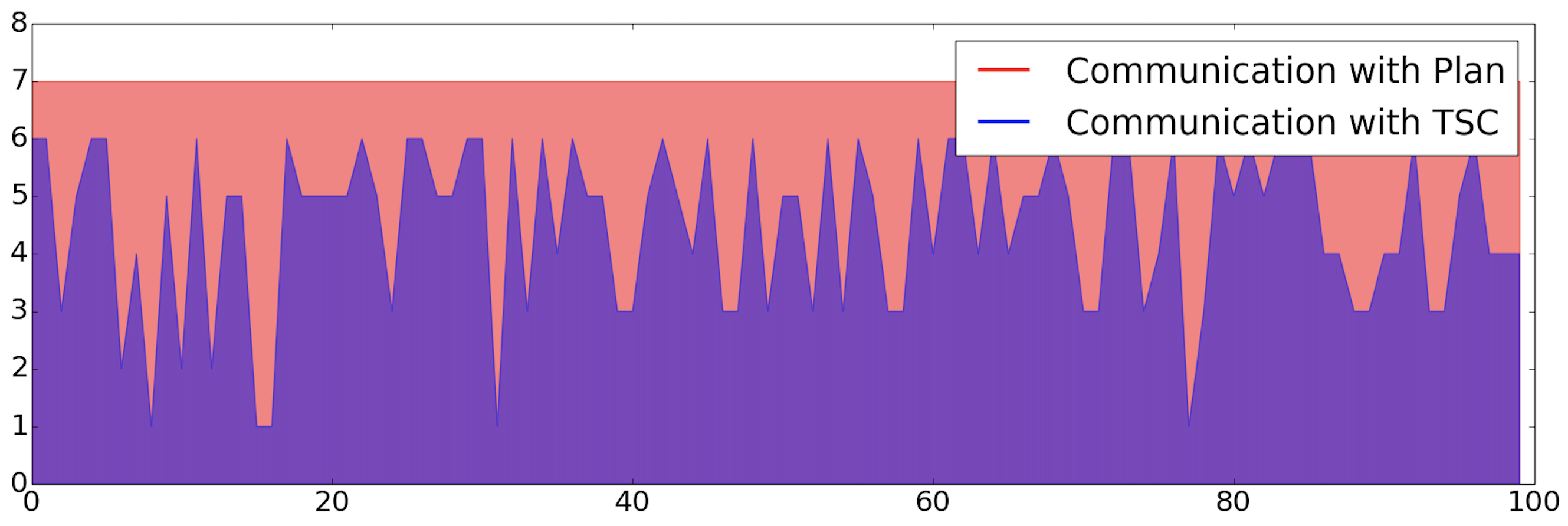}
      \end{subfigure}%
    \hfill
    \begin{subfigure}
        \centering
        \includegraphics[width=0.325\textwidth]{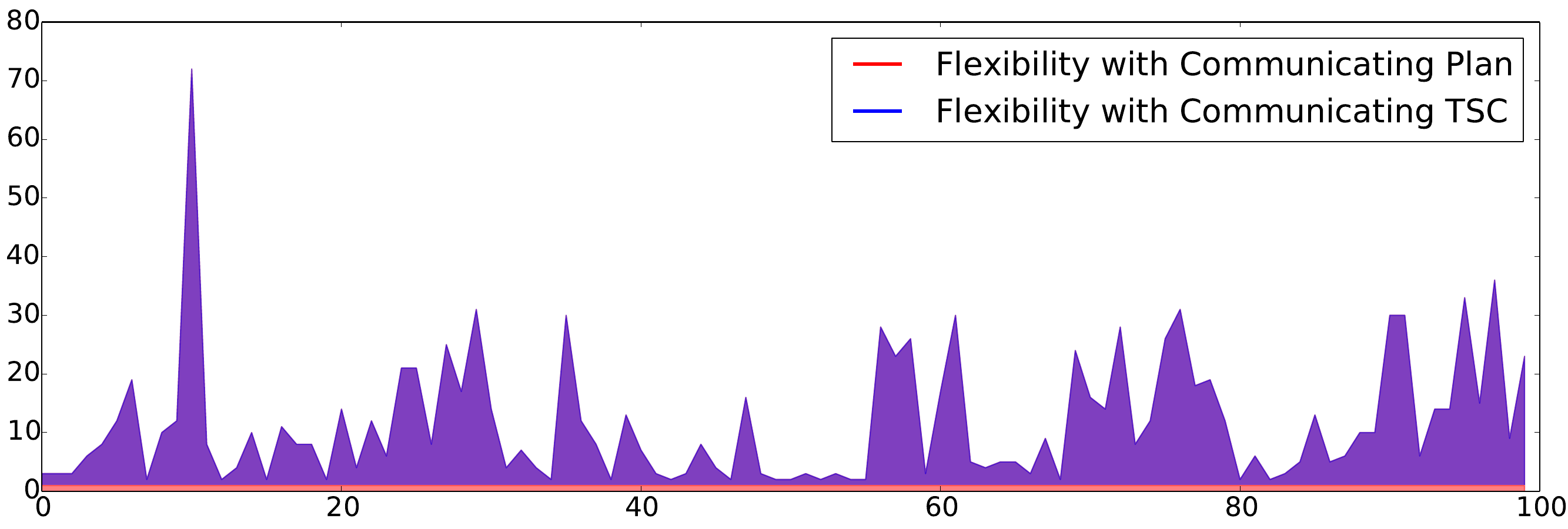}
    \end{subfigure}
    \hfill
    \begin{subfigure}
        \centering
        \includegraphics[width=0.325\textwidth]{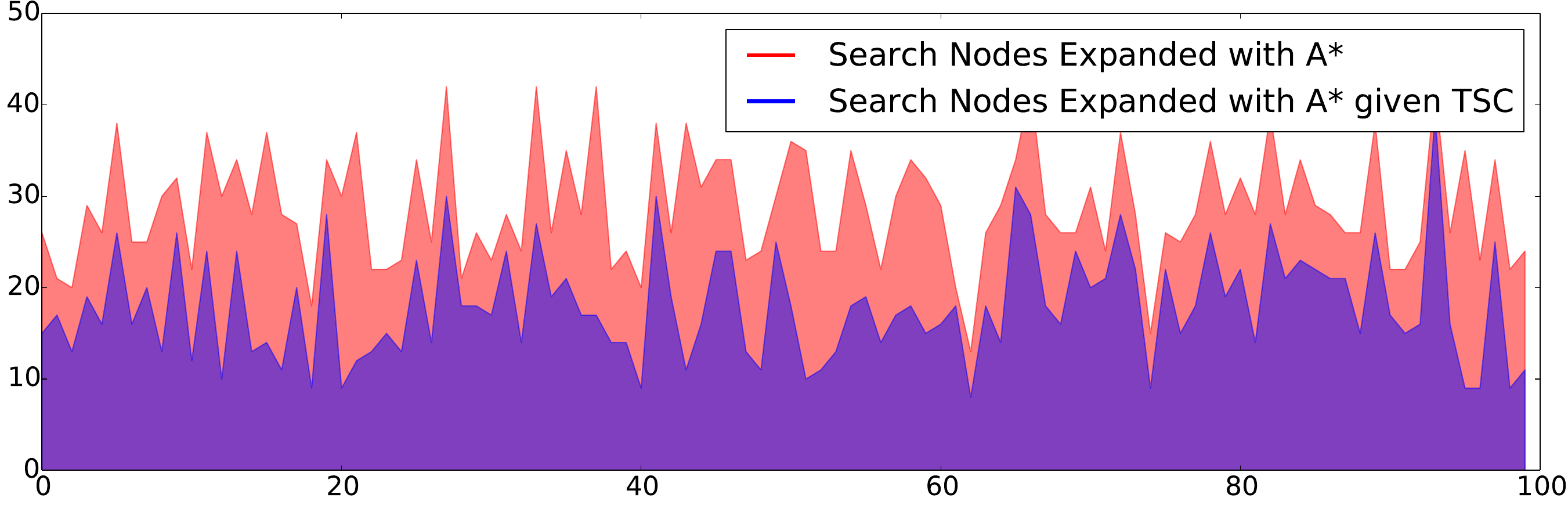}
    \end{subfigure}
        \vskip-5pt
    \caption{\textbf{Left}: Communication cost: plans vs. TSCs for GWB \#7 in Table \ref{tab:abs}, where the state sequence length is always $7$. 
    \textbf{Middle}: Execution flexibility: plans vs. TSCs for GWB $\#7$.
    \textbf{Right}: Planning cost: A* vs. A* with TSC for GWB $\#7$.}
    \label{fig:synthetic}
    \vskip-10pt
\end{figure*}

\begin{figure}[ht]
\vskip-5pt
\centering
{
    \includegraphics[width=0.5\textwidth]{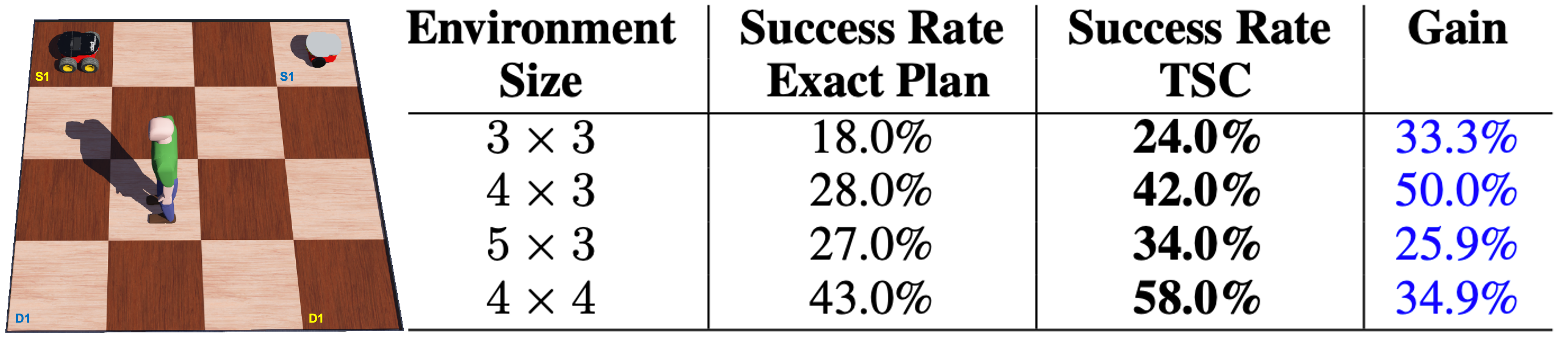}
}
  \vskip-10pt
\caption{{\textbf{Left}: Problem setting for the navigation domain with dynamic obstacles (i.e., a human worker). \textbf{Right}: Success rates of $100$ tasks with dynamic obstacles.}}  
\label{fig:robust}
  \vskip-15pt
\end{figure}

In this evaluation, we implemented the approximate solution for ${D_{MinW^+}}$ in Alg. \ref{alg:alg} (details in supplementary)  
and compared it with the brute-force method discussed in Sec. \ref{subsec:interpreting}. 


\vskip3pt
\noindent \textbf{Language Construction. }
Table \ref{tab:abs} shows the results.
The top part is with open grid-worlds, 
and the bottom with grid-worlds traversable only at the border cells (GWB): all inner cells are untraversable. 
We can see that our approximate solution method is effective in finding coordination languages with a small vocabulary that partitions the state space.
In most cases, our approximate solution returned a vocabulary size that is significantly smaller than the set of ground actions (i.e., $25$ since each robot has $5$ different actions).
This effect is observed even more clearly in the results for GWB, especially when we restricted the set of tasks (i.e., $\mathcal{T}$) by the Manhattan distance between $I$ and $G$ (i.e., the max of the distances between $I$ and $G$ for each robot). 
This constraint is purposely used to remove plans that are too long, and thus difficult to coordinate.
As a side effect, it also removes plans that are too short. 
However, with or without the constraint, we observed that the vocabulary size was more stable in GWB than in GW as the environment size increased, implying the existence of certain features of the coordination languages that were independent of the environment size. 


\vskip3pt
\noindent \textbf{Communication Saving.}
We compared with a baseline that communicates the entire plan. 
The communication cost of the baseline is determined by the length of the plan, and for our method the length of the TSC. 
Given a plan chosen by the communicating agent, 
a sentence is obtained by translating it to the corresponding TSC, which is unique. 
Even though some languages in prior work~\cite{wang2019coordination,wang2019generating} also have this feature, 
translating in the other direction is difficult (see Sec. \ref{sec:rpc}) due to incomplete plan specification~\cite{GINSBERG199589}. 
We chose the setting for GWB \#7 in Table \ref{tab:abs} since the effect of abstraction was more prominent.
 The result is in Fig. \ref{fig:synthetic} (left) for only the first $100/784$ problems where the communication cost differs. 
The average saving is $\textbf{{33.3\%}}$ among the $784$ plans ($\textbf{{27.3\%}}$ among all the $956$ plans). 


\vskip3pt
\noindent \textbf{Increased Flexibility.}
To verify that TSC increases execution flexibility, 
using the same setting as above,
we evaluated the number of plans that were available to a robot (e.g., listener) after receiving a TSC from the other robot (i.e., speaker).
In this evaluation, the speaker always chose the first plan that was found and translated it to a TSC in the coordination language.
The result is in Fig. \ref{fig:synthetic} (middle) for the first $100/774$ problems where the number of available plans with TSC is more than one. 
The increase in flexibility is notable:
the average number of plans is $\textbf{{14.7}}$ among the $774$ plans or  $\textbf{{12.1}}$ among all the $956$ plans (compared to $1$ with the baseline).


\vskip3pt
\noindent \textbf{Reduced Planning Cost.}
\label{sec:rpc}
We evaluated here how coordination could help the listener receiving a TSC reduce its planning cost by providing ``{\it planning guidance}'', compared with planning without such guidance. 
An A* search was implemented with Manhattan distance as the heuristic. 
We also modified the A* search to consider a given TSC (sent by the speaker). 
More specifically, when expanding a node, the modified algorithm would not consider its neighbors that were not aligned with the given TSC. 
The result is presented in Fig. \ref{fig:synthetic} (right) for the first $100$ out of all the $956$ problems. 
The average node reduction is $\textbf{{1.6}}$-fold among all the $956$ problems,
which is substantial.
For the languages studied in~\cite{wang2019coordination,wang2019generating},
the listener would have to compute all the plans and identify one that is compatible with the language expression,
which is more expensive than computing a plan itself!

\subsection{Robust Navigation with Dynamic Obstacles}

Here, we demonstrate how the language 
contributes to robust coordination. 
We consider a warehouse setting (see Fig. \ref{fig:robust} (left)) where robots are tasked to deliver products between one of the storage zones (located at the corners of the workspace and labeled as $S1$ and $S2$) and one of the dispatch zones (located at the other corners of the workspace and labeled as $D1$ and $D2$).  
Products must be transported between the corresponding zones (i.e., $S1-D1$ and $S2-D2$).
For a given task, the robots start randomly from different corners and must deliver, respectively, to the corresponding zones for storage or dispatch. 
At the same time, a human worker may be present in the workspace at a random location other than the four corners. 
We assume that the human worker would not change his/her location during the task. 
Since the robots are from different manufacturers, 
they would not be able to robustly detect each other but can both detect the human. 
To guarantee safety, the robots must coordinate to avoid collisions with each other and the human. 
We assume that robots move at the same speed. 
The robots can coordinate their plans via a coordination language before execution but can only detect the position of the human {\it after} the plan execution starts. 


We tested the success rates of $100$ randomly generated tasks when the robots used the exact plan or a TSC for expressing the plan to coordinate.
When the coordination language is used, the robots can choose other candidate plans (if available) that are expressed by the TSC even when
the initial plan would lead to a collision with the human.
Fig. \ref{fig:robust} (right) shows the results for environments of different sizes where
a language is constructed for each environment. 
We can see that the use of coordination languages significantly improved the success rate in all environments. 
As the environment size increases, the success rates also increased in general since the chance of collision decreases.


\section{Conclusions}
\label{sec:conclusion}
A novel language formation problem was introduced for robust coordination.
We viewed language as a machinery for resolving miscoordinations and
reverse-engineered a language to maximize flexibility during plan execution while guaranteeing optimality. 
We studied the language formation problem under a simplified problem setting
and showed that it is NEXP-complete. 
An approximate solution was then developed and evaluated. 
Our solution, however, still suffers from a scalability issue. 
Our problem shares the same complexity class with finite-horizon DEC-POMDPs so such a limitation should not be surprising and requires future investigation. 
As a result, we keep the evaluation domains minimal for proof-of-concept.
We will publish our code to invite investigation into more practical solutions. 
For more discussions of limitations and applications, refer to supplementary.

\subsubsection{Acknowledgements}
The author would like to thank Li Wang for the initial discussion on this work. 

\bibliographystyle{named}
\bibliography{bib}

\section{Supplementary Materials}

\subsection{Proofs}

\subsubsection{Proof of Theorem 1}
\setcounter{theorem}{0}
\begin{theorem}
The problem  
of deciding whether a coordination language  for domain $M$ under a set of candidate task $\mathcal{T}$ exists with vocabulary $\mathcal{W}$ as a state abstraction of a minimal size $K$ 
and $\mathcal{O} = \{\cdot\}$ (concatenation), or denoted by ${D_{MinW}} = (M, \mathcal{T}, K)$, is NEXP-complete.
\label{NEXP}
\end{theorem}

\begin{proof} 
First, we can solve it by using a Non-deterministic Turing Machine (NTM) in exponential time (without space limit).
This machine can non-deterministically guess $\mathcal{W}$ by enumerating all possible vocabularies under the size of $K$ (including $K$),
which can be done in exponential time in the input size.  
For each candidate vocabulary, checking whether it forms a coordination language with $\mathcal{O} = \{\cdot\}$ will not use more than an exponential amount of time,
which involves computing the set of plans for every candidate task $(I, G) \in \mathcal{T}$ 
and validating that the language relaxes all RCs. 
Next, we show that the decision problem ${D_{MinW}}$ can be reduced from the TILING problem, $D = (L, H, V , N)$, which is known to be NEXP-complete~\cite{papadimitriou1994computational,bernstein2002complexity}.
In a TILING problem, we are given a board of size $N \times N$,
a set of tile types $L = \{tile_0 \dots tile_k\}$, and a set of horizontal and vertical compatibility relations,
$H, V \subseteq L \times L$. 
A tiling is a mapping $p: \{1 \dots N\} \times \{1 \dots N\} \rightarrow L$ and 
is {\textit{{consistent}}} if and only if $p(0, 0) = tile_0$ and each pair of the horizontally or vertically adjacent cells satisfies one of the horizontal or vertical compatibility relations, respectively.
The TILING problem can be reduced to a ${D_{MinW}}$ problem in a two-robot traveling and labeling domain in a grid-world of size $N \times N$ that are 4-way connected.
In this domain, the robots are tasked to move synchronously and label locations consistently with the goal of labeling all locations and visiting all edges. 
The key idea behind the reduction is to let the robots {\it synchronously} discover a consistent tiling via a coordination language. 

The domain $M = (P, A)$ in ${D_{MinW}}$ can be constructed from $D = (L, H, V , N)$ as follows. 
 $P$ includes the propositional state variables for the robots' locations, $u_{x}$.
 Since the robots move synchronously, we only need one location index $x$ in $u_{x}$. 
$P$ also includes variables in the form of $v_{x, l}$, which represents that the location $x$ in the environment is labeled with $l \in L' (L' = L \cup \{l^*\}$). 
It also includes a variable for each (bidirectional) edge to mark whether the edge has been visited, denoted by $m_{x, y}$.
We assume that, initially, the location $(0, 0)$ is labeled with $tile_0$ and all other locations with $l^*$.
Hence, $|P| = N^2 + (N^2 - 1)|L'| + 1 + 2N(N - 1)$. 
For each horizontal relation $(l_1, l_2) \in H$, we introduce $N(N-1)$ pairs of actions, each pair representing the robots synchronously navigating between locations $x$ and $y$ $(x \neq y)$ that are horizontally connected, denoted by $a_{x, y}$ and, reciprocally, $a_{y, x}$.
$a_{x, y}$ has the preconditions of $v_{x, l_1}$, $v_{y, l^*}$, and $u_x$, 
the add effects of $u_{y}$, $v_{y, l_2}$, $m_{x, y}$, 
and the delete effects of $u_x$ and $v_{y, l^*}$. 
For each $(l_1, l_2) \in H$, in addition,
we introduce another $N(N-1)$ pairs of actions, $a_{x, y}$ and, reciprocally, $a_{y, x}$. 
Here, $a_{x, y}$ is similar to the action above except that it now has an additional precondition of $v_{y, l_2}$,
and rids of the precondition of $v_{y, l^*}$, the add effects of $v_{y, l_2}$ and $m_{x, y}$, 
and the delete effect of $v_{y, l^*}$ there. 
We add actions for $V$ in a similar way. 
This results in a total of $2N(N-1) \cdot 2(|H| + |V|)$ actions in $A$.
To set $\mathcal{T}$, we consider only the task of making all $m_{x, y}$ 
present, which are initially absent. 
This is to ensure that all edges must be visited at least once. 
Initially, the robots start at $(0, 0)$. 
A miscoordination occurs when they choose different locations to move to, 
or choose different labels for the same location when it is unlabeled. 
This happens when the robots compute different plans (similar to Fig. 2). 

It can now be shown that the constructed problem of ${D_{MinW}} = (M, \mathcal{T}, 0)$ returns {\it false} if and only if the problem $D$ has a consistent tiling. 
In the first direction, if TILING has a consistent tiling, 
we know that there must exist a traveling plan for the only task considered (i.e., to visit all edges) under $M$.  
This is because the consistent tiling ensures that all locations are connected.
In such a case, the robots, at least, will have to synchronize their first actions (i.e., whether to first perform a horizontal or vertical scan through the environment).
As a result, the vocabulary must have a size of at least $1$. 
In the other direction when no consistent tiling exists, no travel plan can exist 
and hence no coordination is necessary. 
Hence, a coordination language can have a size of $0$ and ${D_{MinW}} = (M, \mathcal{T}, 0)$ would return {\it true}.
\end{proof}

\subsubsection{Proof of Lemma 1}
\setcounter{lemma}{0}
\begin{lemma}
Given a coordination language $(\mathcal{W}, \{\cdot\})$ for domain $M$ under a set of candidate tasks $\mathcal{T}$,
there is another coordination language, $(\mathcal{W}', \{\cdot\})$, where $\mathcal{W}'$ is perfect. 
\label{lem:replace}
\end{lemma}
\begin{proof}
This is easy to see since we can construct a perfect state abstraction $\mathcal{W}'$ from $\mathcal{W}$
by constructing additional abstract states for the overlapping states, and for states not covered by $\mathcal{W}$ in $S$.  
Consider a state space $S = \{1, 2, 3, 4, 5\}$ and a state abstraction  of $\{\{1, 2\}$, $\{2, 3, 4\}\}$.
A perfect state abstraction of $\{\{1\}, \{2\},$ $\{3, 4\}, \{5\}\}$ can be derived accordingly.
For any TSC in the original language, 
a more stringent TSC can be derived using the 
new language with $\mathcal{W}'$.
This new TSC does not violate the condition that the set of expressed plans is RC-free (Def. 4)
since it will only reduce the set of plans expressed. 
It also continues to hold that this new language can express any plan for any candidate task.
\end{proof}

\subsubsection{Proof of Theorem 3}
\setcounter{theorem}{2}
\begin{theorem}
Given the $\mathcal{D}$ above induced from a given domain $M$ and a set of candidate tasks $\mathcal{T}$, if a perfect state abstraction $\mathcal{W}$ (with its mapping denoted by $f^+$) satisfies the condition that
 $\forall (\alpha^*, \beta^*) \in \mathcal{D}$, $\forall i$ $f^+(\alpha^*[i][0]) \neq f^+(\alpha^*[i][1])$ or $\forall i$ $f^+(\beta^*[i][0]) \neq f^+(\beta^*[i][1]) (i \in \{0, 1, 2\})$,
 the language of $(\mathcal{W}, \{\cdot\})$ is a coordination language for $M$ under $\mathcal{T}$. 
 \label{expressive}
\end{theorem}

\begin{proof}
First, it is easy to see that such a language can express any plan given that $f^+$ is a perfect state abstraction, thus
satisfying the {\it completeness} of a coordination language (Def. 4).
Recall that a coordination language must not express any two plans that introduce RC ({\it RC-free}). 
In other words, all plan pairs that introduce RC must {\textit{{always}}} be expressed differently by such a language. 
We show next that a language that satisfies the condition above relaxes the RC for any given plan pair $\pi_1, \pi_2$ and for any $(I, G) \in \mathcal{T}$. 
Given $\pi_1, \pi_2$, 
we can construct a unique TSC $\zeta_1$ and $\zeta_2$ for them, respectively, given a perfect abstraction. 
When the above condition is satisfied, 
we know that $\zeta_1$ and $\zeta_2$ must differ by at least one abstract state and hence
$\pi_1, \pi_2$ will never be expressed by the same TSC (see Fig. 5 for additional explanations). 
As a result, the RC will always be relaxed by the language. 
Since this holds for any plan pair and for any $(I, G) \in \mathcal{T}$, we can conclude that the language is a coordination language for the domain. 
\end{proof}

\subsection{Evaluation}

\subsubsection{Evaluation Domain}

In general, the evaluation domain is similar to the running example. 
At any time step, the robots can move to any adjacent cell or stay. 
The robots are not allowed to stay in the same cell or swap locations at the same time step (to avoid collisions).
While seemingly simple, this setting is in fact challenging and interesting for coordination
since there can be numerous candidate plans for a given task. 
For example, given that only the makespan is considered, while waiting for the other robot to reach its goal, a robot may wander around, stay at its goal, or temporarily stay at any intermediate locations.
The results were obtained with an implementation in Java on a MacBook Pro with Intel Core  i7 CPU 2.80GHz and 16GB of RAM.

In our evaluation, to find all plans, we used a modified A* by allowing the search to continue after a goal is found and terminate at any suboptimal solution. 
To further improve the approximate solution, 
we considered only $\alpha^*$ in Theorem \ref{expressive} and implemented a greedy algorithm to determine the minimal state abstraction, i.e., adding any remaining states to an abstract state until no more can be added. 

\subsubsection{Evaluation in Coordination Game}

\setcounter{figure}{8}
\begin{figure}[h!]
\centering
{
    \includegraphics[width=0.37\textwidth]{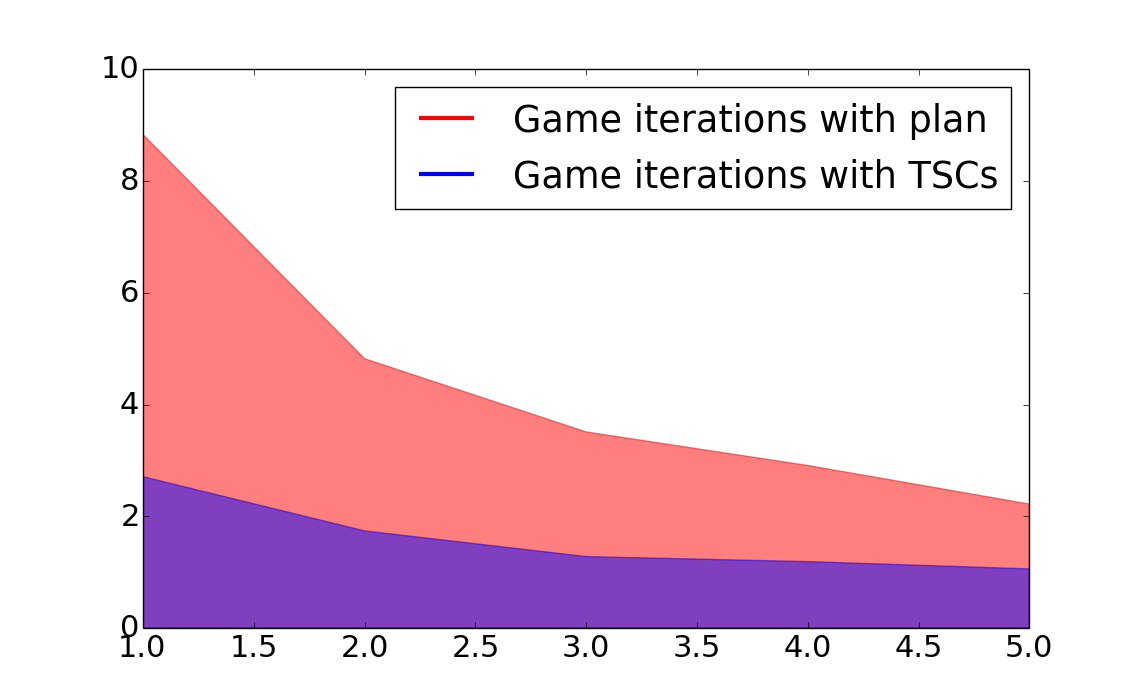}
}
\caption{Comparison of the number of iterations played by the speaker and listener between communicating plans and communicating TSCs for GWB $\#5$ in Table 1. The X-axis shows the $k$ value and the Y-axis the number of iterations played until the speaker's proposal is accepted by the listener.
The results are averaged over $100$ runs.  
}
\label{fig:game}
\vskip-10pt
\end{figure}

In this evaluation, we verify the benefits of the flexibility provided by the language during negotiation in a coordination game. 
In this game, 
the robots are given a random task to achieve.
The speaker first computes its plan (which is implemented by randomly choosing a candidate plan for the task). 
At the same time, we assume that the listener randomly chooses  
 $k$ different plans as its preferences 
out of all the candidate plans for the task.
The speaker then acts as a proposer that either communicates its plan or the TSC to the listener. 
The listener would only accept the proposal when the speaker's request is compatible with any of its preferred plans,
and in which case the game ends.
Otherwise, the game continues from the beginning (the speaker recomputes its plan and listener reselects its preferences) until the proposal of the speaker is eventually accepted. 
We are interested to see how many iterations the robots must play this game when
plans and TSCs are communicated by the speaker, respectively. 
We ran this game on GWB \#5 described in Table 1. 
The results are averaged over $100$ runs as $k$ is changed from $1$ to $5$, and are shown in Fig. \ref{fig:game}.
We can see that the number of iterations decreases as $k$ increases in both cases, which is expected.
However, these numbers remain relatively consistent and much smaller in the TSC case, which clearly illustrates the benefit of TSCs for reducing negotiation costs. 



\subsection{Discussions}

Coordinating multiple agents can be either achieved by considering joint states of all agents for language construction or separately coordinating each agent pair via the language constructed for $2$ agents. 
The former is expensive while the latter's communication cost grows polynomially with the number of agents, making it best suitable for loosely-coupled domains. 
Furthermore, the latter cannot consider miscoordinations that occur only with more than two agents.
The problem setting considered is simplified in many aspects. 
For example, in multi-agent planning, the agents often do not have access to the full joint domain model.
It would be interesting to consider partial models that may differ from the perspectives of different robots. 
The problem to coordinate robots under such model differences is a significant challenge. 
Generalizing the language more specifically to robots of different types should also be considered,
as well as when continuous domains are concerned. 
Our work ignored coordination via observations and actions to focus on explicit coordination. 
However, observations and actions can greatly simplify coordination  since we can use them to resolve a subset of miscoordinations without communication. 
However, incorporating actions and observations in coordination would require a more complex machinery, e.g., 
considering language formation in a Dec-POMDP setting. 

The guarantee for optimality may be impractical and unnecessary in some cases. 
Extending our language definition to allow sub-optimality would further increase its flexibility and practicality. 
However, this will likely increase the number of candidate plans, 
leading to more miscoordinations. 
To address this issue, 
we can consider {\it probabilistically miscoordination-free} to relax the language requirement for more possibilities, similar to that in~\cite{GINSBERG199589}.
This would result in a spectrum of languages that vary in complexities and performance guarantees. 

Application-wise, robots communicating ``{\it plan sketches}'' may also be used as a way to conserve privacy~\cite{brafman2015privacy}.
The flexibility in choosing plans may alternatively be interpreted as hiding information (e.g., plan preferences) from other agents.
It would be interesting to study the type of information that can be protected. 
Another possible application of our language is to consider communication denied environments
so that the language can be designed to handle situations where certain parts of the environment are more prone to communication challenges.  
In such cases, coordination may be facilitated before entering and after exiting those parts of the environment by using the proposed languages to maximize the robustness in between. 

\subsubsection{Relation to Prior work}
It is worth noting that the temporal-state specification of a TSC may be considered as a strict generalization of options in semi-MDPs~\cite{SUTTON1999181},
which are specified by the initiation and termination conditions.  
Instead, a TSC specifies a sequence of abstract states that must be visited by a plan that satisfies it in order.  
While this distinction may not be important for planning purposes since a TSC can be constructed from options,
it could affect the construction of a language substantially since TSCs allow for more flexible expressions.  
On the other hand, TSCs represent a subset of constraints that can be specified by LTL and CTL~\cite{vardi1996automata,emerson1990temporal}.

Recent work has studied similar problems under the pretext of language formation~\cite{wang2019coordination,wang2019generating}. 
These methods, however, are more related to the idea of approximate planning~\cite{GINSBERG199589}.
Although we draw inspiration from these prior works that also use plan space abstraction,  
 the focus there is 
 to specify {partial plans}, 
which are {\it incomplete specifications} of plans.
Our language, on the other hand, is designed to specify a set of complete plans.
One of the key limitations of prior work is that there is no easy way to translate between partial plans and complete plans (at least in one direction). 
It implies that, for example, the partial plans cannot be used to guide planning during coordination as desired (see Sec. 4).

\end{document}